\documentclass[review]{elsarticle}
\usepackage{geometry}
\usepackage{comment}

\usepackage{amsmath,amssymb,amscd,latexsym,dsfont,wasysym,bbm}
\usepackage{float}
\usepackage{color}
\usepackage{graphicx}
\usepackage{multirow,multicol} 
\usepackage{verbatim}
\usepackage{psfrag}
\usepackage{comment}
\usepackage{makeidx}

\usepackage[frozencache,cachedir=mintedcache]{minted}
\usepackage[]{algorithm}
\usepackage{algorithmicx}
\usepackage{algpseudocode}
\algnewcommand\algorithmicinput{\textbf{Initialize:}}
\algnewcommand\Initialize{\item[\algorithmicinput]}

\usepackage{cases}
\usepackage{setspace}
\usepackage{pgfplots}
\newcommand{\tr}[1]{\textrm{#1}}
\newcommand{\mr}[1]{\mathrm{#1}}
\newcommand{\tnr}[1]{{\textnormal{#1}}}

\newcommand{\mf}[1]{\mathsf{#1}}
 
\newcommand{\ms}[1]{\mathds{#1}}

\newcommand{\ov}[1]{\overline{#1}}

\newcommand{\bb}{\boldsymbol{b}}

\newcommand{\bd}{\boldsymbol{d}}

\newcommand{\be}{\boldsymbol{e}}

\newcommand{\bg}{\boldsymbol{g}}

\newcommand{\bh}{\boldsymbol{h}}

\newcommand{\br}{\boldsymbol{r}}

\newcommand{\bs}{\boldsymbol{s}}

\newcommand{\bu}{\boldsymbol{u}}

\newcommand{\bv}{\boldsymbol{v}}
\newcommand{\bw}{\boldsymbol{w}}

\newcommand{\bx}{\boldsymbol{x}}

\newcommand{\by}{\boldsymbol{y}}

\newcommand{\bzero}{\boldsymbol{0}}

\newcommand{\blambda}{\boldsymbol{\lambda}}

\newcommand{\bsigma}{\boldsymbol{\sigma}}

\newcommand{\bphi}{\boldsymbol{\phi}}

\newcommand{\ie}{i.e.,~} 		
\newcommand{\eg}{e.g.,~}	

\newcommand{\argmin}{\mathop{\mr{argmin}}}

\newcommand{\set}[1]{\{#1\}}

\newcommand{\ld}{\ldots}

\newcommand{\e}{\mr{e}}

\newcommand{\diag}[1]{\mathop{\mathrm{diag}\left(#1\right)}\nolimits}

\newcommand{\Ex}{\ms{E}}     			
\newcommand{\T}{^{\top}}            		
\newcommand{\Th}{^{\mr{H}}}             		
\newcommand{\dd}{\,\mr{d}}             		

\newcommand{\mfm}{\mf{m}}

\newcommand{\Real}{\mathbb{R}}		
\newcommand{\Complex}{\mathbb{C}}		

\newcommand{\matA}{\tnr{\textbf{A}}}
\newcommand{\matB}{\tnr{\textbf{B}}}

\newcommand{\matD}{\tnr{\textbf{D}}}

\newcommand{\matF}{\tnr{\textbf{F}}}
\newcommand{\matG}{\tnr{\textbf{G}}}
\newcommand{\matH}{\tnr{\textbf{H}}}
\newcommand{\matI}{\tnr{\textbf{I}}}

\newcommand{\matK}{\tnr{\textbf{K}}}

\newcommand{\matQ}{\tnr{\textbf{Q}}}
\newcommand{\matR}{\tnr{\textbf{R}}}
\newcommand{\matS}{\tnr{\textbf{S}}}

\newcommand{\matU}{\tnr{\textbf{U}}}
\newcommand{\matV}{\tnr{\textbf{V}}}
\newcommand{\matW}{\tnr{\textbf{W}}}
\newcommand{\matX}{\tnr{\textbf{X}}}

\newcommand{\matZ}{\tnr{\textbf{Z}}}

\usepackage{mathtools}
\usepackage[only,llbracket,rrbracket]{stmaryrd}

\newenvironment{aligneq}
{\begin{equation}
\begin{aligned}
}
{
\end{aligned} 
\end{equation}
}

\usepackage{tikz}
\pgfplotsset{compat=1.18}

\usetikzlibrary{
    arrows.meta,
    automata,
    positioning,
    shadows,
    calc,
    math,
    shapes.geometric,
    decorations,
    decorations.pathmorphing,
    decorations.pathreplacing,
    decorations.shapes,
    decorations.markings,
    graphs
}

\tikzset{%
dot/.style={
  circle,
  inner sep=0mm,
  outer sep=0mm,
  minimum size=2mm,
  draw=black,
  fill=black
},
triple/.style={
  double distance=2pt,
  postaction={draw}
},
quadruple/.style={
  double,
  double distance=2pt,
  postaction={
    draw,
    transform canvas={yshift=-.4pt},
  },
  postaction={
    draw,
    transform canvas={yshift=.4pt},
  }
},
every loop/.style={},
d/.style={
 Circle[]-,
 shorten <=-2pt,
 transform canvas={shift={(-3pt, 4pt)}}
},
d0/.style={
 Circle[]-,
 shorten <=-2pt,
},
d1/.style={
 Circle[]-,
},
dn/.style={
 draw, circle, minimum size=1.3em
},
ddn/.style={
 draw, circle, double, minimum size=1.3em
},
dddn/.style={
 draw, circle, double, minimum size=1.3em,
   double distance=2pt,
   postaction={draw}
},
triangle/.style={
regular polygon,
regular polygon sides=3,
rotate=270,
scale=.5,
inner sep=3pt,
draw
},
}

\usepackage[acronym,nonumberlist]{glossaries} 
\usepackage{hyperref}
\usepackage{empheq}
\usepackage{verbatim}
\usepackage{graphicx}
\usepackage{subfig}
\newacronym[\glsshortpluralkey=PDFs,\glslongpluralkey=probability density functions]{pdf}{PDF}{probability density function}
\newacronym[\glsshortpluralkey=CDFs,\glslongpluralkey=cumulative density functions]{cdf}{CDF}{cumulative density function}
\newacronym[\glsshortpluralkey=CCDFs,\glslongpluralkey=complementary cumulative density functions]{ccdf}{CDF}{complementary cumulative density function}
\newacronym[\glsshortpluralkey=PMFs,\glslongpluralkey=probability mass functions]{pmf}{PMF}{probability mass function}
\newacronym[]{lhs}{l.h.s.}{left-hand side}
\newacronym[]{rhs}{r.h.s.}{right-hand side} 

\newacronym[]{bicm}{BICM}{bit-interleaved coded modulation}
\newacronym[]{bicmid}{BICM-ID}{BICM with iterative demapping}
\newacronym[]{cm}{CM}{coded modulation}
\newacronym[]{tcm}{TCM}{trellis-coded modulation}
\newacronym[]{mlc}{MLC}{multi-level coding}
\newacronym[]{pam}{PAM}{pulse amplitude modulation}
\newacronym[]{bpsk}{BPSK}{binary phase shift keying}
\newacronym[]{qam}{QAM}{quadrature amplitude modulation}
\newacronym[]{16qam}{16-QAM}{16-points quadrature amplitude modulation}
\newacronym[]{psk}{PSK}{phase shift keying}
\newacronym[\glsshortpluralkey=LLRs,\glslongpluralkey=logarithmic likelihood ratios]{llr}{LLR}{logarithmic likelihood ratio}
\newacronym[]{oc}{OC}{operating characteristic}
\newacronym[]{dmp}{DMP}{discretized message passing}
\newacronym[]{mp}{MP}{message passing}
\newacronym[]{ep}{EP}{expectation propagation}
\newacronym[\glsshortpluralkey=MIs,\glslongpluralkey=mutual informations]{mi}{MI}{mutual information}
\newacronym[\glsshortpluralkey=GMIs,\glslongpluralkey=generalized mutual informations]{gmi}{GMI}{generalized mutual information}
\newacronym[]{eesm}{EESM}{exponential effective-SNR-mapping}
\newacronym[]{bicm-gmi}{BICM-GMI}{BICM generalized mutual information}
\newacronym[]{awgn}{AWGN}{additive white Gaussian noise}
\newacronym[]{bsc}{BSC}{binary symetric channel}
\newacronym[]{amc}{AMC}{adaptive modulation and coding}
\newacronym[]{csi}{CSI}{channel state information}
\newacronym[]{cqi}{CQI}{channel quality indicator}
\newacronym[]{kl}{KL}{Kullback-Leibler}
\newacronym[]{cmm}{CMM}{circular moment matching}
\newacronym[]{ga}{GA}{Gaussian approximation}

\newacronym[]{sp}{SP}{set-partitioning}
\newacronym[]{gsm}{GSM}{global system for mobile communications}
\newacronym[]{edge}{EDGE}{enhanced data rates for GSM evolution}
\newacronym[]{3gpp}{3GPP}{3rd generation partnership project}
\newacronym[]{umts}{UMTS}{Universal Mobile Telecommunication System}
\newacronym[]{lte}{LTE}{Long Term Evolution}
\newacronym[]{dvb}{DVB}{digital video broadcasting}
\newacronym[]{fdd}{FDD}{Frequency Division Duplexing}

\newacronym[\glsshortpluralkey=CCs,\glslongpluralkey=convolutional codes]{cc}{CC}{convolutional code}
\newacronym[\glsshortpluralkey=PCCCs,\glslongpluralkey=parallel concatenated convolutional codes]{pccc}{PCCC}{parallel concatenated convolutional code}
\newacronym[\glsshortpluralkey=TCs,\glslongpluralkey=turbo codes]{tc}{TC}{turbo code}
\newacronym{ldpc}{LDPC}{low-density parity-check}
\newacronym[]{ofdm}{OFDM}{orthogonal frequency-division multiplexing}
\newacronym[]{bep}{BEP}{bit-error probability}
\newacronym[]{wep}{WEP}{word-error probability}
\newacronym[]{sep}{SEP}{symbol-error probability}
\newacronym[]{pep}{PEP}{pairwise-error probability}
\newacronym[]{ttcm}{TTCM}{turbo-trellis coded modulation}
\newacronym[]{uep}{UEP}{unequal error protection}
\newacronym[\glsshortpluralkey=CENCs,\glslongpluralkey=convolutional encoders]{cenc}{CENC}{convolutional encoder}
\newacronym[]{mimo}{MIMO}{multiple-input multiple-output}
\newacronym[\glsshortpluralkey=SNRs,\glslongpluralkey=signal-to-noise ratios]{snr}{SNR}{signal-to-noise ratio}
\newacronym[\glsshortpluralkey=SINRs,\glslongpluralkey=the signal-to-interference-plus-noise ratios]{sinr}{SINR}{the signal-to-interference-plus-noise ratio}
\newacronym[]{msb}{MSB}{most-significative bit}
\newacronym[]{bcjr}{BCJR}{Bahl--Cocke--Jelinek--Raviv}
\newacronym[]{cbc}{CBC}{Colavolpe--Barbieri--Caire}
\newacronym[]{skr}{SKR}{Shayovitz--Kreimer--Raphaeli}
\newacronym[\glsshortpluralkey=SEDs,\glslongpluralkey=squared Euclidean distances]{sed}{SED}{squared Euclidean distance}
\newacronym[\glsshortpluralkey=EDs,\glslongpluralkey=Euclidean distances]{ed}{ED}{Euclidean distance}
\newacronym[\glsshortpluralkey=MEDs,\glslongpluralkey=minimum Euclidean distances]{med}{MED}{minimum Euclidean distance}
\newacronym[]{core}{CoRe}{constellation rearrangement}
\newacronym[]{msd}{MSD}{multistage decoding}
\newacronym[]{pdl}{PDL}{parallel decoding of the individual levels}
\newacronym[\glsshortpluralkey=GCs,\glslongpluralkey=Gray codes]{gc}{GC}{Gray code}
\newacronym[]{brgc}{BRGC}{binary-reflected Gray code}
\newacronym[]{nbc}{NBC}{natural binary code}
\newacronym[]{fbc}{FBC}{folded-binary code}
\newacronym[]{bsgc}{BSGC}{binary semi-Gray code}
\newacronym[]{msp}{MSP}{modified set-partitioning}
\newacronym[]{ssp}{SSP}{semi set-partitioning}
\newacronym[]{fhd}{FHD}{free Hamming distance}
\newacronym[]{mfhd}{MFHD}{maximum free Hamming distance}
\newacronym[]{ods}{ODS}{optimal distance spectrum}
\newacronym[]{iud}{i.u.d.}{independent and uniformly distributed}
\newacronym[]{ud}{u.d.}{uniformly distributed}
\newacronym[]{iid}{i.i.d.}{independent, identically distributed}
\newacronym[]{ami}{AMI}{accumulated mutual information}
\newacronym[]{bico}{BICO}{binary-input continuous-output}
\newacronym[]{gh}{GH}{Gauss--Hermite}
\newacronym[]{gum}{GUM}{Gaussian--uniform mixture}

\newacronym[\glsshortpluralkey=BSs,\glslongpluralkey=base-stations]{bs}{BS}{base-station}
\newacronym[\glsshortpluralkey=MSs,\glslongpluralkey=mobile-stations]{ms}{MS}{mobile-stations}

\newacronym[]{phy}{PHY}{physical layer} 
\newacronym[]{rlc}{RLC}{Radio-Link control} 
\newacronym[]{ran}{RAN}{Radio Access Network} 
\newacronym[]{llc}{LLC}{logical link control} 
\newacronym[]{tcp}{TCP}{transmission control protocol} 
\newacronym[]{mac}{MAC}{media access control} 
\newacronym[]{fft}{FFT}{fast Fourier transform} 
\newacronym[]{ft}{FT}{Fourrier transform}
\newacronym[]{cf}{CF}{characteristic function} 
\newacronym[]{mgf}{MGF}{moment generating function} 
\newacronym[]{ee}{EE}{energy efficiency} 
\newacronym[]{eb}{EB}{energy per bit}
\newacronym[]{kkt}{KKT}{Karush--Kuhn--Tucker} 
\newacronym[]{mcs}{MCS}{modulation/coding scheme} 
\newacronym[]{fec}{FEC}{forward error correction}
\newacronym[]{arq}{ARQ}{automatic repeat request}
\newacronym[]{harq}{HARQ}{hybrid ARQ}
\newacronym[]{tarq}{TARQ}{truncated HARQ}
\newacronym[]{ir}{IR}{incremental redundancy}
\newacronym[]{rpr}{RR}{repetition redundancy}
\newacronym[]{rrharq}{RR-HARQ}{repetition redundancy HARQ}
\newacronym[]{irharq}{IR-HARQ}{incremental redundancy HARQ}
\newacronym[]{ack}{ACK}{positive acknowledgment}
\newacronym[]{nack}{NACK}{negative acknowledgment}
\newacronym[]{hol}{HoL}{head of the line}
\newacronym[]{crc}{CRC}{cyclic redundancy check}
\newacronym[]{dp}{DP}{dynamic programming}
\newacronym[]{gp}{GP}{geometric programming}
\newacronym[]{per}{PER}{packet error rate}
\newacronym[]{ber}{BER}{bit error rate}
\newacronym[]{op}{OP}{outage probability}
\newacronym[]{spa}{SPA}{saddle-point approximation}
\newacronym[]{mrc}{MRC}{maximum ratio combining}
\newacronym[]{mdp}{MDP}{Markov decision process}
\newacronym[]{lp}{LP}{linear programming}
\newacronym[]{pomdp}{POMDP}{partially observable Markov decision process}
\newacronym[]{psimdp}{PSI-MDP}{partial state information Markov decision process}
\newacronym[]{scpp}{SCPP}{stochastic shortest path problem}

\newacronym[]{forw}{frwd}{forward}
\newacronym[]{feed}{fdbk}{feedback}

\newacronym[]{mm}{MM-HARQ}{multi-message HARQ}
\newacronym[]{xp}{XP-HARQ}{cross-packet HARQ}
\newacronym[]{ts}{TS}{time-sharing}
\newacronym[]{sc}{SC}{superposition coding}
\newacronym[]{sbrq}{SBRQ}{systematic backward retransmission}
\newacronym[]{brq}{BRQ}{backward retransmission}
\newacronym[]{lharq}{L-HARQ}{layer-coded HARQ}
\newacronym[]{anlharq}{AoN-HARQ}{all-or-none L-HARQ}
\newacronym[]{vlharq}{VL-HARQ}{variable-length HARQ}

\newacronym[]{pp}{PPP}{point process}
\newacronym[]{ppp}{PPP}{Poisson point process}

\newacronym[]{fide}{FIDE}{F\'ed\'eration Internationale des \'Echecs}
\newacronym[]{fifa}{FIFA}{F\'ed\'eration Internationale de Football Association}
\newacronym[]{fivb}{FIVB}{F\'ed\'eration Internationale de Volleyball}
\newacronym[]{epl}{EPL}{English Premier League}
\newacronym[]{nhl}{NHL}{National Hockey League}
\newacronym[]{nfl}{NFL}{National Football League}

\newacronym[]{sg}{SG}{stochastic gradient}
\newacronym[]{lms}{LMS}{least mean squares}
\newacronym[]{rls}{RLS}{recursive least squares}
\newacronym[]{vss}{VSS}{variable step-size}
\newacronym[]{hfa}{HFA}{home-field advantage}
\newacronym[]{ha}{HA}{home advantage}
\newacronym[]{mov}{MOV}{margin of victory}
\newacronym[]{ac}{AC}{Adjacent Categories}
\newacronym[]{cl}{CL}{Cumulative Link}
\newacronym[]{rps}{RPS}{Ranked Probability Score}
\newacronym[]{mse}{MSE}{Mean Square Error}
\newacronym[]{mmse}{MMSE}{Minimum Mean Square Error}
\newacronym[]{rmse}{RMSE}{Root Mean Squares Error}
\newacronym[]{map}{MAP}{maximum a posteriori}
\newacronym[]{ml}{ML}{maximum likelihood}
\newacronym[]{loo}{LOO}{leave-one-out}
\newacronym[]{alo}{ALO}{approximate leave-one-out}

\newacronym[]{svd}{SVD}{singular values decomposition}

\newacronym[]{skf}{SKF}{Simplified Kalman Filter}
\newacronym[]{vskf}{vSKF}{\emph{vector-covariance} Simplified Kalman Filter}
\newacronym[]{sskf}{sSKF}{\emph{scalar-covariance} Simplified Kalman Filter}
\newacronym[]{fskf}{fSKF}{\emph{fixed-variance} Simplified Kalman Filter}
\newacronym[]{kf}{KF}{Kalman Filter}
\newacronym[]{gelo}{G-Elo}{Generalized Elo}

\newacronym[]{tpb}{TPB}{tensor-product-basis}

\usepackage[svgnames]{xcolor}
\definecolor{cblue}{HTML}{1965B0}

\definecolor{cred}{HTML}{B8221E}

\definecolor{dgreen}{rgb}{0,0.6,0}

\definecolor{dorange}{RGB}{255, 128, 0}

\definecolor{burntorange}{rgb}{0.8, 0.33, 0.0}

\newacronym[]{cp}{CP}{Canonical Polyadic}
\newacronym[]{ols}{OLS}{ordinary least squares}

\begin{document}

\begin{frontmatter}

\title{Automatic regularization for Kronecker product MMSE filters}


\author[1]{Daniel~Gomes~de~Pinho~Zanco}
\ead{daniel.zanco@inrs.ca}

\author[1]{Leszek~Szczecinski}
\ead{leszek.szczecinski@inrs.ca}

\author[1]{Jacob~Benesty}
\ead{jacob.benesty@inrs.ca}

\author[2]{Eduardo~Vinicius~Kuhn}
\ead{kuhn@utfpr.edu.br}

\address[1]{INRS–Institut National de la Recherche  Scientifique, Montreal, QC, H5A-1K6, Canada.}

\address[2]{Federal University of Technology - Paran\'a, Toledo, Paran\'a, 85902-490, Brazil.}


\begin{abstract}

In this work, we propose a method to automatically find the regularization parameter for low-rank MMSE filters based on a Kronecker-product representation, using an approximate leave-one-out (ALO) cross-validation metric. We formulate the problem in a constrained form that covers both error minimization and interference suppression, and explore the link between regularization and rank selection. The proposed method is validated on system identification and antenna beamforming problems, showing significant gains over commonly used methods.


\end{abstract}

\begin{keyword} Kronecker product representation, low-rank system, MMSE filter, regularization, system identification, beamforming.



\end{keyword}

\end{frontmatter}

\section{Introduction}\label{sec:intro}


Linear minimum mean-squared error (MMSE)  filters play a central role in a wide range of signal processing tasks, including channel equalization \cite[Ch.~5.4]{Sayed2008}, system identification \cite{Dogariu2021}, antenna beamforming \cite[Ch.~6.5]{Sayed2008}, among others. Estimating these filters typically requires an implicit or explicit inversion of the input signal's covariance matrix, which, to ensure numerical stability and a well-defined solution, may require \emph{regularization}. The latter can be achieved in different ways \cite{Dogariu2021}, \eg by penalizing the filter parameters with an energy constraint, \ie by introducing a positive regularization term to the diagonal of the covariance matrix; or by enforcing a particular structure of the filter, such as a low-rank form.

Low-rank models are also important in machine learning due to their ability to capture the essence of complex data while reducing its dimensionality \cite{Udell2016}. They are especially interesting for regression problems, since these models offer significant advantages over complete or high-dimensional models in terms of overfitting reduction  \cite[Ch.~7]{Hastie2009}, which is particularly important in scenarios where only a small amount of data is available. 

Previous results obtained in various signal processing applications \cite{Benesty2019,Kuhn2021,Dogariu2021,He2023} indicate the enormous potential of low-rank Kronecker modeling, which consists in representing the MMSE filter as the Kronecker product\footnote{To avoid confusion, as a naming convention, we use the term ``Kronecker product'' to denote the modeling principle, where a filter is represented by the product of vectors, which must be estimated from the data. This approach is different from the ``Kronecker decomposition'', which is concerned with representing a known (or already estimated) parameter (here, filter) with a low-rank structure.} of a set of smaller filters, and show that the adaptation can be performed with less data, reducing overfitting. 



Despite the relevance of regularization and its widespread usage, the appropriate tuning of its hyperparameters is often neglected, which is the case in the literature of tensor decomposition \cite{Kolda2009,Ballard2025} and tensor regression \cite{Liu2021}. In the signal processing literature, \cite{GomesdePinhoZanco2025} shows how the regularization parameter can be obtained from the data via a simple closed-form fixed-point iteration.



Some approaches to address regularization exist also in tensor decomposition, \eg \cite{Navasca2008} uses an iterative Tikhonov regularization, where the parameter decays after each iteration, but no hyperparameter search is done for the regularization. In the matrix and tensor regression literature, \cite{Zhou2013a} uses the Akaike information criterion (AIC) and Bayesian information criterion (BIC) as metrics for parameter selection in nuclear-norm regularized matrix regression, while \cite{Guhaniyogi2017} applies Bayesian modeling, but since it requires distribution sampling, it is computationally expensive. \cite{Wang2024} addresses robust sparse representation through a combination of sparse and low-rank regularizations on the residual; its regularization parameters are ``adaptively adjusted during iterations'', although no update rule is provided. \cite{Sun2025} addresses tensor-structured wireless channel estimation through a Bayesian formulation, in which regularization is implicit in the prior and the hyperparameters are obtained by evidence maximization.

In this work, we
\begin{enumerate}
\item propose a method for finding the regularization parameter in low-rank MMSE filters based on the Kronecker-product representation;
\item formulate the problem in a constrained form covering both error minimization and interference suppression, and extend the ALO derivation to this constrained case;
\item explore the theoretical connection between regularization and rank, and the importance of the appropriate choice of the regularization parameter in low-rank models; and
\item compare our approach to other regularization strategies, on experiments covering both system identification and beamforming.
\end{enumerate}



This MMSE filtering problem is formulated in Sec.~\ref{sec:problem}, where its low-rank counterpart is presented together with our main theoretical contribution. Next, in Sec.~\ref{sec:als}, we present the algorithm used to solve it and show an efficient cross-validation method for finding the regularization parameter. In Sec.~\ref{sec:results}, we apply the proposed method to system identification and antenna beamforming problems and demonstrate its effectiveness by comparing it to other methods, including an unattainable in practice ``oracle'' solution. In Sec.~\ref{sec:conclusion} we present our main conclusions.



\section{Problem definition}\label{sec:problem}


We consider the constrained linear filtering problem, where an input signal $\bx(n) \in \Complex^M$ is linearly combined with/filtered by the weights $\bw \in \Complex^M$, producing the output signal $y(n) \in \Complex$. 
Consider the following MMSE problem
\begin{aligneq}
\label{eq:problem}
\hat\bw & = \argmin_{\bw}  \Ex\left[|y(n) - \bw\Th\bx(n)|^2\right] \quad \tnr{s. t. }\quad \bd\Th\bw = c\\
&= \ov\matR_{\bx}^{-1}(\ov{\br}_{\bx y} - \ov\lambda \bd),
\end{aligneq}
where $c \in \Complex$ is a constraint scalar, $\bd$ is the response generated by the signal $y(n)$, with $\|\bd\|^2=M$, $\Ex[\cdot]$ is the expectation calculated with respect to $\bx(n)$ and $y(n)$, $\ov\matR_{\bx} = \Ex\left[\bx(n)\bx\Th(n)\right]$, $\ov{\br}_{\bx y} = \Ex\left[\bx(n) y^*(n)\right]$, and
\begin{align}
\ov\lambda = \frac{\bd\Th\ov\matR_{\bx}^{-1}\ov{\br}_{\bx y} - c}{\bd\Th\ov\matR_{\bx}^{-1}\bd}
\end{align}
is the Lagrange multiplier. The problem \eqref{eq:problem} is found in system identification \cite{Dogariu2021}, channel equalization \cite[Ch.~2]{Haykin2014}, antenna beamforming \cite{Kuhn2021}, and many others.

In practice, we do not have access to $\ov\matR_{\bx}$ or $\ov\br_{\bx y}$. Instead, we estimate them from the data using time-averaging, i.e.,
\begin{align}\label{eq:left.Rx}
\ov\matR_{\bx} &\approx \matR_{\bx}=\frac{1}{N} \sum_{n=0}^{N-1}\bx(n)\bx\Th(n),\\
\label{eq:left.rxd}
\ov{\br}_{\bx y} &\approx \br_{\bx y} =\frac{1}{N} \sum_{n=0}^{N-1}\bx(n)y^*(n).
\end{align}
Replacing $\ov\matR_{\bx}$ with $\matR_{\bx}$ and $\ov{\br}_{\bx y}$ with $\br_{\bx y}$ in \eqref{eq:problem} yields the \emph{empirical} version of the MMSE equation. When the number of available samples $N$ is small relative to the number of weights $M$, time-averaging may introduce random errors in these empirical measures, and to ensure numerical stability when $\matR_{\bx}$ is ill-conditioned, the problem is typically regularized. Regularization introduces penalties that impose structure to the solution, thus leading to a simpler model, which avoids overfitting, as well as introduces robustness to errors in the specification of the constraint $\bd$. This leads to the empirical formulation of the problem:
\begin{align}
\label{unified.problem}
\hat\bw & = \argmin_{\bw}  \frac{1}{N} \| \by - \matX\Th\bw\|_2^2 + \alpha r(\bw) \quad \tnr{s. t. }\quad \bd\Th\bw = c,
\end{align}
where $\matX = \big[\bx(0),\ld,\bx(N-1)\big]\T \in \Complex^{M \times N}$, $\by = \big[y^*(0),\ld,y^*(N-1)\big]\T \in \Complex^N$, $\alpha \geq 0$ is the regularization parameter, $r(\bw): \Complex^M \rightarrow \Real$ is a regularization function, and $\| \cdot \|_2$ is the Euclidean norm.

The standard choice for $r(\bw)$ is the squared $\ell_2$-norm penalty (Tikhonov/ridge regularization), which yields
\begin{align}
\hat\bw 
\label{eq:ridge}
&= \argmin_{\bw}  \frac{1}{N} \| \by - \matX\Th\bw\|_2^2 + \alpha \|\bw\|_2^2 \quad \tnr{s. t. }\quad \bd\Th\bw = c\\
\label{l2.solution}
&= (\matR_{\bx} + \alpha \matI)^{-1}({\br}_{\bx y} - \lambda \bd),
\end{align}
in which $\matI$ is the identity matrix, and the Lagrange multiplier
\begin{align}
\lambda = \frac{\bd\Th(\matR_{\bx} + \alpha \matI)^{-1}{\br}_{\bx y} - c}{\bd\Th(\matR_{\bx} + \alpha \matI)^{-1}\bd}.
\end{align}

An alternative regularizer, used to promote sparsity in $\hat\bw$, is the $\ell_1$ regularization (lasso) \cite{Tibshirani1996}
\begin{align}
\hat\bw 
\label{eq:lasso}
&= \argmin_{\bw}  \frac{1}{N} \| \by - \matX\Th\bw\|_2^2 + \alpha \|\bw\|_1 \quad \tnr{s. t. }\quad \bd\Th\bw = c
\end{align}
which has no closed-form solution but can be solved efficiently as well.

Regularization deals with limited data and helps us control the form of the solution: $\ell_2$ shrinks all coefficients of $\hat\bw$ continuously towards zero, while $\ell_1$ sets coefficients to exactly zero past a parameter-dependent threshold, producing a sparse solution. Thus, the choice of the regularizer $r(\bw)$ has to take into account the properties of the \emph{solution} $\hat\bw$, and is not a structural reduction of the estimation problem itself: \eg the optimization \eqref{eq:lasso} is still over the full $\bw\in\Complex^M$, and only the data determines which coefficients ultimately vanish. Nevertheless, sparsity is desirable because, even if computationally we must deal with $M$ parameters, only a (small) subset of them improves the quality of the estimation.

\subsection{Low-rank via Kronecker product}\label{sec:lowrank}


A different way to regularize the solution is to reduce, \emph{structurally}, the number of free parameters in the optimization itself. The MMSE solution \eqref{l2.solution} treats $\bw$ as an unstructured vector of $M$ free parameters, and if $M$ is large relative to $N$, this leads to overfitting. When dealing with 2D signals, we have instead a parameter matrix $\matW \in \Complex^{M_1 \times M_2}$, with $M = M_1 M_2$, of which we can penalize its rank, \ie the number of non-zero singular values of $\matW$. The 2D structure may emerge from different practical settings, even when the original objects are not necessarily 2D: \eg acoustic impulse responses with repeating patterns; antenna arrays, where the steering vector is exactly separable; models with separable covariance; or space–time/MIMO filters, which are naturally 2D.



As in the case of the lasso solution \eqref{eq:lasso} promoting sparseness in the parameters, here we may take the $\ell_1$-norm of the singular values to promote a low-rank structure, leading to the problem
\begin{equation}
\begin{aligned}
\label{cp.nuclear.reg}
\hat\matW 
=& \argmin_{\matU^{(1)}, \matU^{(2)}} \frac{1}{N}\| \by - \matX\Th\tr{vec}(\matW)\|^2_2 + 2\alpha \|\matW\|_{*}\\
&\quad \tnr{s. t. } \quad \bd\Th\tr{vec}(\matW) = c,
\end{aligned}
\end{equation}
where $\|\matW\|_* = \sum_{i=1}^{\min(M_1,M_2)}\sigma_i(\matW)= \|\bsigma(\matW)\|_1$ is the nuclear norm of $\matW$ and $\sigma_i(\matW)$ are its singular values.

In practice, using the nuclear norm complicates the optimization, as it is non-differentiable and its evaluation requires an SVD. Rather, it is preferable to use the variational characterization of the nuclear norm \cite[Appendix~A.1]{Scarvelis2024}, \cite[Appendix~A.5]{Mazumder2010},
\begin{align}\label{eq:variational.nuclear}
\|\matW\|_* = \min_{\matW = \matU^{(1)}(\matU^{(2)})\T} \frac{1}{2}\left(\|\matU^{(1)}\|_\tr{F}^2+\|\matU^{(2)}\|_\tr{F}^2\right),
\end{align}

where we express $\matW$ as a sum of $R$ outer products,
\begin{equation}\label{cp.w}
\begin{aligned}
\matW &= \sum_{r=1}^R \bu_{r}^{(1)} \circ \bu_{r}^{(2)}
= \matU^{(1)} \left(\matU^{(2)} \right)\T\in\Complex^{M_1\times M_2},
\end{aligned}
\end{equation}
where $\circ$ denotes the outer product, and the factor matrices
\begin{align}\label{factor.definition}
\matU^{(k)} = [\bu_1^{(k)}, \bu_2^{(k)}, \ldots, \bu_R^{(k)}] \in \Complex^{M_k \times R} , \quad k=1,2,
\end{align}
in which $\bu_{r}^{(1)} \in \Complex^{M_1}$, $\bu_{r}^{(2)} \in \Complex^{M_2}$, and $R$ is the \emph{construction rank} of $\matW$ and satisfies $R\ge\tnr{rank}(\matW)$.

In the Kronecker-product filtering literature \cite{Benesty2021,He2023}, the decomposition \eqref{cp.w} is meant to reduce the number of elements to be estimated; consequently, the construction rank $R\leq\min\set{M_1, M_2}$ controls that number and becomes a hyperparameter to be tuned alongside $\alpha$. Finding $R$ thus becomes a problem of \emph{rank selection}, which the literature often treats as a regularization/modelling problem. The formulation above makes this unnecessary. These findings are well-known in related research that uses matrix norms or its variational counterparts to control rank in other types of problems \cite{Xie2017,Gong2019,Zhang2024,Zhang2024a,Zhang2024b}, but in these works the regularization parameter is either set heuristically or via small grid searches; an estimation method is not provided, which is what we aim to address later in Sec.~\ref{sec:auto.reg}.

The problem \eqref{cp.nuclear.reg} can then be expressed in the factor domain,
\begin{equation}
\begin{aligned}
\label{cp.reg.regression}
\hat\matU^{(1)}, \hat\matU^{(2)} =& \argmin_{\matU^{(1)}, \matU^{(2)}} \frac{1}{N}\|\by - \matX\Th\tr{vec}(\matW)\|^2_2 + \alpha\left( \big\|\matU^{(1)}\big\|_{\tr{F}}^2 + \big\|\matU^{(2)}\big\|_{\tr{F}}^2 \right) \\
&\quad \tnr{s. t. } \ \matW = \matU^{(1)}  \left(\matU^{(2)} \right)\T, \quad \bd\Th\tr{vec}(\matW) = c,
\end{aligned}
\end{equation}
where each factor is regularized by its squared Frobenius norm, with estimate
\begin{align}\label{w.hat.matrix}
\hat\matW = \hat\matU^{(1)} \left(\hat\matU^{(2)} \right)\T.
\end{align}
Note that, by rescaling the factors $\matU^{(1)} \to a\,\matU^{(1)}$, $\matU^{(2)} \to a^{-1}\matU^{(2)}$, which leaves $\matW$ invariant, the more general formulation with two regularization parameters $\alpha_1, \alpha_2$ collapses to the single-parameter form \eqref{cp.reg.regression} with $\alpha = \sqrt{\alpha_1\alpha_2}$.




Although the equivalence \eqref{cp.reg.regression}--\eqref{cp.nuclear.reg} is itself well-known, it reframes the problem and interpretation of the solution: in \eqref{cp.reg.regression} the penalty looks like an extension of ridge regression, but the penalty in \eqref{cp.nuclear.reg} is an $\ell_1$-norm acting on the singular values $\bsigma(\matW)$. \emph{Regularization in the low-rank Kronecker-product MMSE problem is thus lasso in the spectral domain of $\matW$}, even though no $\ell_1$ penalty is explicitly imposed on the optimization variables $\matU^{(1)},\matU^{(2)}$. This perspective has direct consequences that mirror standard lasso behavior in the coefficient domain \cite{Tibshirani1996,Shang2021}:
\begin{itemize}
\item \emph{Discrete rank reduction along the regularization path.} The rank of $\hat\matW(\alpha)$ is a piecewise-constant, non-increasing function of $\alpha$, with rank reductions occurring at specific values of $\alpha$ at which singular values are driven to zero. This is analogous to how the support of a lasso estimate is piecewise constant in $\alpha$, with coefficients deactivating at discrete events along the path.
\item \emph{Existence of $\alpha_{\max}$.} In the unconstrained case ($c=0$, $\bd=\bzero$), there exists a finite threshold $\alpha_{\max}$ such that, for all $\alpha\ge\alpha_{\max}$, the solution satisfies $\hat\matW(\alpha)=\mathbf{0}$, i.e., all singular values are zero. This is the spectral counterpart of the well-known $\alpha_{\max}$ of lasso. An explicit formula and proof for the nuclear-norm-regularized regression problem are given in \cite[Lemma~3.1, Theorem~3.1]{Shang2021}, which corresponds to
\begin{align}\label{alpha.max}
\alpha_{\max} = \sigma_{\max}[\tr{mat}_{M_1,M_2}(\br_{\bx y})],
\end{align}
where $\tr{mat}_{K,L}: \Complex^{KL} \rightarrow \Complex^{K \times L}$ is a matricization operator, and $\sigma_{\max}(\cdot)$ is the maximum singular value of a given matrix. For the constrained case, $\alpha_{\max}$ may not exist, depending on the constraint given.


\end{itemize}

A practical implication is that the construction rank $R$ in \eqref{cp.w} need not be tuned as a hyperparameter. Trying to ``select'' $R$ in fact imposes an additional burden, because each candidate $R$ requires its own choice of regularization parameter $\alpha$. It suffices instead to choose $R$ large enough not to constrain the solution---e.g., $R=\min\set{M_1,M_2}$---and let $\alpha$ control the effective rank of $\hat\matW$ via the implicit nuclear-norm penalty in \eqref{cp.nuclear.reg}. This replaces the burden of selecting $R$ with the one of choosing the right regularization parameter $\alpha$; this motivates the data-driven selection method we develop in Section~\ref{sec:auto.reg}.

\subsection{Alternating Least Squares}\label{sec:als}

The main method to solve problem \eqref{cp.reg.regression} is called alternating minimization (AM) [or block coordinate descent (BCD)] \cite[Ch.~3.7]{Bertsekas2016}. In our case, the AM is called alternating least squares (ALS), as it consists of solving a least-squares problem for each factor while keeping the other constant, \ie for $k=1,2$,
\begin{equation}\label{als.iter.K}
\begin{aligned}
\hat\bu^{(k)} \leftarrow& \argmin_{\bu^{(k)}} \frac{1}{N}\| \by - \matX\Th \hat\matA^{(k)} \bu^{(k)}\|^2_2 + \alpha \|\bu^{(k)}\|^2_2\\
&\tr{s. t. } \bd\Th \hat\matA^{(k)} \bu^{(k)} = c,
\end{aligned}
\end{equation}
where $\hat\bu^{(k)} = \tr{vec}\left(\hat\matU^{(k)} \right) \in \Complex^{M_k R}$, the matrices $\hat\matA^{(k)} \in \Complex^{M \times M_k R}$ are such that
\begin{align}
\hat\matA^{(1)}\hat\bu^{(1)} = \hat\matA^{(2)}\hat\bu^{(2)} = \tr{vec}\left[\hat\matU^{(1)}  \left(\hat\matU^{(2)} \right)\T\right],
\end{align}
with the matrices $\hat\matA^{(1)}$ and $\hat\matA^{(2)}$ being obtained from
\begin{align}
\label{eq:A.hat.1}
\hat\matA^{(1)} &= \matA^{(1)}\left(\hat\matU^{(2)}\right) \in \Complex^{M \times M_1 R},\\
\label{eq:A.hat.2}
\hat\matA^{(2)} &= \matA^{(2)}\left(\hat\matU^{(1)}\right) \in \Complex^{M \times M_2 R},
\end{align}
in which $\matA^{(1)}$ and $\matA^{(2)}$ are functions of the factors $\matU^{(2)}$ and $\matU^{(1)}$, respectively, and are defined in \ref{app:factors}. 

The solution of the sub-problems (for $k=1,2$) is given by
\begin{align}
\label{hat.bu.k}
\hat\bu^{(k)} &= \left( \matR_{\bx}^{(k)} + \alpha \matI \right)^{-1} \left( \br_{\bx y}^{(k)} - \lambda^{(k)} \bd^{(k)} \right),
\end{align}
where
\begin{align}
\matR_{\bx}^{(k)} &= \left(\hat\matA^{(k)}\right)\Th \matR_{\bx} \hat\matA^{(k)} \in \Complex^{M_k R \times M_k R},\\
\br_{\bx y}^{(k)} &= \left(\hat\matA^{(k)}\right)\Th \br_{\bx y} \in \Complex^{M_k R},\\
\bd^{(k)} &= \left(\hat\matA^{(k)}\right)\Th \bd,\\
\lambda^{(k)} &= \frac{\bd^{(k) \mr{H}}(\matR_{\bx}^{(k)} + \alpha \matI)^{-1}{\br}_{\bx y}^{(k)} - c}{\bd^{(k) \mr{H}}(\matR_{\bx}^{(k)} + \alpha \matI)^{-1}\bd^{(k)}}.
\end{align}
Therefore, the ALS algorithm can be described using the pseudo-code in Algorithm \ref{alg:als}, in which $I_{\tr{ALS}}$ denotes the number of iterations. Note that the convergence of BCD/ALS is guaranteed when each sub-problem \eqref{als.iter.K} has a unique minimizer (see \cite[Prop.~3.7.1]{Bertsekas2016} for a proof), which is true when $\alpha > 0$.
\begin{algorithm}
\caption{Alternating Least Squares (ALS)}\label{alg:als}
\begin{algorithmic}[1]
\Require $I_{\tr{ALS}}$
\Initialize $\alpha, 
\matU_0^{(2)}$
\State 
$\hat{\bu}^{(2)} \gets 
\tr{vec}\left(\matU_0^{(2)}\right)$
\For{$i=1$ to $I_{\tr{ALS}}$}
\For{$k=1$ to $2$}
\State Compute $\hat\matA^{(k)}$ with \eqref{eq:A.hat.1} or \eqref{eq:A.hat.2}
\State $\matR_{\bx}^{(k)} \gets \left(\hat\matA^{(k)}\right)\Th \matR_{\bx} \hat\matA^{(k)}$
\State $\br_{\bx y}^{(k)} \gets \left(\hat\matA^{(k)}\right)\Th \br_{\bx y}$
\State $\bd^{(k)} \gets \left(\hat\matA^{(k)}\right)\Th \bd$
\State $\lambda^{(k)} \gets \displaystyle\frac{\bd^{(k) \mr{H}}(\matR_{\bx}^{(k)} + \alpha \matI)^{-1}{\br}_{\bx y}^{(k)} - c}{\bd^{(k) \mr{H}}(\matR_{\bx}^{(k)} + \alpha \matI)^{-1}\bd^{(k)}}$
\State $\hat\bu^{(k)} \gets \left( \matR_{\bx}^{(k)} + \alpha \matI \right)^{-1} \left( \br_{\bx y}^{(k)} - \lambda^{(k)} \bd^{(k)} \right)$
\EndFor
\EndFor

\Return $\hat\bu^{(1)}, \hat\bu^{(2)}$
\end{algorithmic}
\end{algorithm}

\textbf{Initialization via decomposition:} A commonly used initialization in tensor decomposition consists of performing high-order singular value decomposition (HOSVD) \cite{Ballard2025,Kolda2009} on the tensor in question and then truncating to the desired rank. For the 2-dimensional case, HOSVD reduces to SVD. Nevertheless, in the Kronecker-product least-squares problem \eqref{cp.reg.regression}, we do not have access to the true filter parameters $\matW$. Instead, we can use the full-rank solution \eqref{l2.solution} as a starting point, and take its rank-$R$ approximation:
\begin{align}
\tr{mat}_{M_1,M_2}(\hat\bw) &\approx \matQ \tr{diag}(\bs) \matV\Th,
\end{align}
where $\matQ \in \Complex^{M_1 \times R}$, $\matV \in \Complex^{M_2 \times R}$, $\bs \in \Real_+^{R}$ constitute the truncated SVD of $\tr{mat}_{M_1,M_2}(\hat\bw)$. So, using the SVD matrices, we obtain the initialization factors:
\begin{align}
\matU^{(1)}_0 &= \tr{diag}(\sqrt{\bs}) \matQ,\\
\matU^{(2)}_0 &= \tr{diag}(\sqrt{\bs}) \matV^*,
\end{align}
where $\sqrt{\cdot}$ operates point-wise on elements of $\bs$. Note that we only need $\matU^{(2)}_0$ to initialize the ALS algorithm. 


\section{Automatic regularization}\label{sec:auto.reg}

We want to address the issue of finding the regularization parameter $\alpha$ from the data. This issue remains largely unexplored in the literature on tensor decomposition \cite{Ballard2025,Cichocki2016} and regression \cite{Liu2021,Li2018}, and it is rarely addressed in the area of signal processing. However, it is relatively well-known in the machine learning (ML) literature and, recently, the ML methods have also been adopted to find regularization in the context of linear MMSE filters \cite{GomesdePinhoZanco2025,Selen2008,Ledoit2004}. In this section, we propose a method to find the regularization parameter $\alpha$ in the design of low-rank MMSE filters based on \eqref{cp.reg.regression}, where, due to the Kronecker-product structure of the filter, conventional methods are not applicable, as there's no linearity in the parameters. We derive a method in the following.

\subsection{Leave-one-out cross-validation}

A conventional method in ML for evaluating the performance of a model is the use of cross-validation \cite[Ch.~7.10]{Hastie2009}. It consists of removing (leaving out) part of the training samples from the optimization and computing the performance over the removed data; this is done over multiple left-out parts, over which the performance is averaged. When we remove one sample, optimize over the remaining samples, and repeat the procedure over all samples to average the performance, this is known as leave-one-out (LO) cross-validation. 

In the case of Kronecker-product optimization \eqref{cp.reg.regression}, LO cross-validation is defined as
\begin{align}
\label{eq:loo.func}
J_{\tr{LO}}(\alpha) 
= \frac{1}{N} \sum_{n=1}^N \left| y_n - \bx_n\Th\tr{vec}\big(\hat\matW_{\backslash n}\big) \right|^2,
\end{align}
where
\begin{align}
\hat\matW_{\backslash n} &= \hat\matU_{\backslash n}^{(1)}  \left(\hat\matU_{\backslash n}^{(2)} \right)\T,
\end{align}
and
\begin{equation}
\label{eq:loo.kron}
\begin{aligned}
\hat\matU_{\backslash n}^{(1)}, \hat\matU_{\backslash n}^{(2)} =& \argmin_{\matU^{(1)}, \matU^{(2)}} \frac{1}{N}\big\|\by_{\backslash n} - \matX_{\backslash n}\Th\tr{vec}(\matW)\big\|^2_2 + \alpha \sum_{k=1}^2 \big\|\matU^{(k)}\big\|_{\tr{F}}^2 \\
&\quad \tr{s. t. } \ \matW = \matU^{(1)}  \left(\matU^{(2)} \right)\T, \quad \bd\Th\tr{vec}(\matW) = c,
\end{aligned}
\end{equation}
with the subscript $n$ denoting the index of the removed sample, while $\by_{\backslash n} \in \Complex^{N-1}$ and $\matX_{\backslash n} \in \Complex^{M \times (N-1)}$ correspond to $\by$ and $\matX$ with the $n$th sample removed.

However, implementing \eqref{eq:loo.kron} directly requires solving it $N$ times, which is infeasible for large $N$. Thus, simplifications must be sought.

Some validation methods are often considered in place of LO, such as $K$-fold cross-validation \cite[Ch.~7.10]{Hastie2009}, where the average in \eqref{eq:loo.func} is done over $K$ sets instead; or a simple out-of-sample validation set, which splits out only one part of the training samples (essentially a $1$-fold cross-validation). Although these approaches may reduce computation time, they are affected by the arbitrary choice of the sets and can lead to inaccuracies in the cross-validation metric (see \cite[Fig.~1]{Rad2020a} for an intuitive illustration of the $K$-fold cross-validation metric).

\subsection{Approximate leave-one-out cross-validation}

An interesting alternative is the ALO cross-validation \cite{Rad2020a}, which consists in obtaining a closed-form expression that approximates the LO solution by applying one step of Newton's method to the LO problem, initialized at the full-sample estimate.\footnote{Note that if the loss function is quadratic, the LO solution can be obtained exactly as is the case in the so-called PRESS method \cite{Allen1971}, which solves the LO cross-validation for the full-rank problem \eqref{eq:ridge}.} However, since ALO considers an unconstrained problem where the loss function is linear in the parameters (see \cite[Eq.~1]{Rad2020a}), we cannot naively apply it to \eqref{cp.reg.regression}. Thus we proceed with a derivation of \emph{an} ALO-like metric for the problem in question.




We start by redefining \eqref{eq:loo.kron} as
\begin{equation}
\label{eq:loo.kron.vec}
\hat\bu_{\backslash n} = \argmin_{\bu} L_{\backslash n}(\bu) \quad \tr{s. t. }  C(\bu) = 0,
\end{equation}
where
\begin{align}
L_{\backslash n}(\bu) &= L(\bu) - \ell_n(\bu),\\
C(\bu) &= \frac{1}{2}\bd\Th\matA(\bu)\bu - c,
\end{align}
in which
\begin{align}
\bu &= 
\begin{bmatrix}
\bu^{(1)\top} & \bu^{(2)\top}
\end{bmatrix}\T \in \Complex^P,\\
\matA(\bu) &= 
\begin{bmatrix}
\matA^{(1)}\left(\bu^{(2)}\right) & \matA^{(2)}\left(\bu^{(1)}\right)
\end{bmatrix}
\in \Complex^{M \times P},
\end{align}
with $P = (M_1 + M_2) R$ and recalling the definition of $\matA^{(1)}$ and $\matA^{(2)}$ [see \eqref{eq:A.1} and \eqref{eq:A.2}, respectively], and
\begin{align}\label{eq:obj.fun}
L(\bu) &= \sum_{i=1}^N \ell_i(\bu) + N\alpha \left\|\bu\right\|^2_2,
\end{align}
where
\begin{aligneq}
\ell_i(\bu) 
&= \left| y_i - \frac{1}{2}\bx_i\Th\matA(\bu)\bu \right|^2,
\end{aligneq}
with $\bu_{\backslash n} = [\bu_{\backslash n}^{(1)\top}, \bu_{\backslash n}^{(2)\top}]\T \in \Complex^P$ and $\hat\bu^{(k)}_{\backslash n} = \tr{vec} \left( \hat\matU^{(k)}_{\backslash n} \right)$ for $k=1,2$. Note that, in contrast to vanilla ALO \cite{Rad2020a}, the problem \eqref{eq:loo.kron.vec} is constrained [by $C(\bu) = 0$], thus requiring a constrained Newton's method.

We then approximate $\hat\bu_{\backslash n}$ by one step of Newton's method, initialized at $\hat\bu$,
\begin{align}\label{eq:newton.update}
\hat\bu_{\backslash n} \approx \tilde\bu_{\backslash n} = \hat\bu + \tilde\bv_{\backslash n},
\end{align}
where $\tilde\bv$ is obtained by minimizing, over the feasible set, the second-order Taylor-series approximation of $L_{\backslash n}(\bu)$ centered around $\hat\bu$, \ie
\begin{aligneq}
\label{eq:newton.step}
\tilde\bv_{\backslash n} = \argmin_{\bv} \;& 2\tr{Re}\{\nabla_{\bu^*} L_{\backslash n} (\hat\bu)\Th \bv\} + \bv\Th\nabla_{\bu^*}^2 L_{\backslash n}(\hat\bu)\bv + \tr{Re}\{\bv\T\nabla_{\bu}^2 L_{\backslash n}(\hat\bu)\bv \}\\
&\tr{ s. t. } C(\hat\bu + \bv) = 0,
\end{aligneq}
where $\hat\bu = [\hat\bu^{(1)\top}, \hat\bu^{(2)\top}]\T \in \Complex^P$ is the solution obtained using Algorithm \ref{alg:als}, and
\begin{align}
\label{diff.1st}
\nabla_{\bu^*}L_{\backslash n}(\bu) &= \frac{\dd L_{\backslash n}}{\dd\bu\Th}\biggr\rvert_{\bu = \hat\bu},\\
\label{diff.2nd.herm}
\nabla_{\bu^*}^2 L_{\backslash n}(\bu) &= \frac{\dd^2 L_{\backslash n}}{\dd\bu\dd\bu\Th}\biggr\rvert_{\bu = \hat\bu},\\
\label{diff.2nd.real}
\nabla^2_{\bu}L_{\backslash n}(\bu) &= \frac{\dd^2 L_{\backslash n}}{\dd\bu\dd\bu\T} \biggr\rvert_{\bu = \hat\bu}.
\end{align}

In order to solve \eqref{eq:newton.step}, we have to compute the first and second order derivatives of $L_{\backslash n}(\bu)$, which is bilinear in the parameters [due to $\matA(\bu)$], as well as find the linear approximation to the constraint $C(\bu)$ (see \ref{app:alo} for the derivation).

First, linearizing the constraint $C(\bu)$ around $\hat\bu$ leads to
\begin{align}
\label{eq:constraint.linearized}
C(\hat\bu + \bv) \approx \bd\Th\hat\matA\bv,
\end{align}
where $\hat\matA = \matA(\hat\bu)$.

For the first order derivative \eqref{diff.1st}, since $\hat\bu$ minimizes $L(\bu)$ subject to $C(\bu) = 0$, this implies $\nabla_{\bu^*} L(\hat\bu) = -\hat\lambda \hat\matA\Th\bd$, for some Lagrange multiplier $\hat\lambda$. Evaluating at $\hat\bu$ then gives
\begin{aligneq}
\label{eq:grad.alo}
\nabla_{\bu^*} L_{\backslash n}(\hat\bu) &= \nabla_{\bu^*} L(\hat\bu) - \nabla_{\bu^*} \ell_n(\hat\bu)\\
&= -\hat\lambda\hat\matA\Th\bd + \hat\matA\Th\bx_n\hat{e}_n,
\end{aligneq}
where
\begin{align}\label{e.hat.n}
\hat{e}_n = y_n - \bx_n\Th\hat\bw
\end{align}
and
\begin{align}
\hat\bw = \frac{1}{2} \hat\matA\hat\bu = \hat\matA^{(k)}\hat\bu^{(k)}, \quad k=1,2.
\end{align}

And, for the second order derivatives \eqref{diff.2nd.herm} and \eqref{diff.2nd.real}, we have, respectively,
\begin{aligneq}
\label{eq:hess.alo}
\nabla^2_{\bu^*} L_{\backslash n}(\hat\bu)
&= \matF - \hat\matA\Th\bx_n\bx_n\Th\hat\matA,
\end{aligneq}
where
\begin{aligneq}\label{mat.F}
\matF &= \sum_{i=1}^N \hat\matA\Th\bx_i\bx_i\Th\hat\matA + N\alpha\matI\\
&= N \hat\matA\Th\matR_{\bx}\hat\matA + N\alpha\matI,
\end{aligneq}
and
\begin{aligneq}
\label{eq:hess.real.alo.pre}
\nabla^2_{\bu} L_{\backslash n}(\hat\bu)
&= -\sum_{i \neq n}^N \hat\matB_i,
\end{aligneq}
where
\begin{align}
\hat\matB_i &=
\begin{bmatrix}
\bzero & \matI_R \otimes \matS_i\\
\matI_R \otimes \matS_i\T & \bzero
\end{bmatrix},
\end{align}
in which $\matS_i = \tr{mat}_{M_1, M_2}(\hat{e}_i^*\bx_i^*)$. We then note that, from the orthogonality principle \cite[Ch.~2.2]{Haykin2014}, $\Ex[\hat{e}_n\bx_n] = \bzero$, then
\begin{align}
\label{eq:mean.Cn}
\sum_{i \neq n}^N \hat\matB_i \approx \bzero,
\end{align}
thus we may ignore the last term in \eqref{eq:newton.step}.


Applying \eqref{eq:constraint.linearized}, \eqref{eq:grad.alo}, and \eqref{eq:hess.alo} into \eqref{eq:newton.step}, we then express the Newton step as
\begin{aligneq}
\label{eq:newton.step.explicit}
\tilde\bv_{\backslash n} = \argmin_{\bv} \;& \bv\Th\left(\matF - \hat\matA\Th\bx_n\bx_n\Th\hat\matA\right)\bv + 2\tr{Re}\{\hat{e}_n\bx_n\Th \hat\matA \bv\}\\
&\tr{ s. t. } \bd\Th\hat\matA\bv = 0,
\end{aligneq}
in which the term $-\hat\lambda\hat\matA\Th\bd$ from the gradient \eqref{eq:grad.alo} is removed since it is in the feasible set and contributes nothing to the constrained
step. The problem can be solved using the same method as \eqref{l2.solution}, yielding
\begin{align}\label{eq:newton.step.solution}
\tilde\bv_{\backslash n} = -\hat{e}_n\left(\matF - \hat\matA\Th\bx_n\bx_n\Th\hat\matA\right)^{-1}\hat\matA\Th( \bx_n - \lambda_n\bd),
\end{align}
where
\begin{align}\label{eq:lagrange.n}
\lambda_n = \frac{\bd\Th\hat\matA\left(\matF - \hat\matA\Th\bx_n\bx_n\Th\hat\matA\right)^{-1}\hat\matA\Th\bx_n}{\bd\Th\hat\matA\left(\matF - \hat\matA\Th\bx_n\bx_n\Th\hat\matA\right)^{-1}\hat\matA\Th\bd}.
\end{align}

Now, using the Sherman-Morrison inversion lemma \cite[Sec.~3.2.4]{Petersen2012}, we have
\begin{align}\label{inversion.lemma}
\left(\matF - \hat\matA\Th\bx_n\bx_n\Th\hat\matA\right)^{-1} = \matF^{-1} + \frac{\matF^{-1}\hat\matA\Th\bx_n\bx_n\Th\hat\matA\matF^{-1}}{1 - \bx_n\Th\hat\matA\matF^{-1}\hat\matA\Th\bx_n},
\end{align}
which we use to express \eqref{eq:newton.step.solution} as
\begin{aligneq}\label{eq:newton.step.v}
\tilde\bv_{\backslash n}
&= -\frac{\hat{e}_n}{1 - z_n} \,\bg_n 
\end{aligneq}
where
\begin{align}
\label{gn}
\bg_n &= \matF^{-1}\hat\matA\Th\bx_n - 
\underbrace{\displaystyle\frac{\bd\Th\hat\matA\matF^{-1}\hat\matA\Th\bx_n}{\bd\Th\hat\matA\matF^{-1}\hat\matA\Th\bd}\matF^{-1}\hat\matA\Th\bd}_\tr{constraint correction},\\
\label{zn}
z_n &= \bx_n\Th\hat\matA\bg_n =  \bx_n\Th\hat\matA\matF^{-1}\hat\matA\Th\bx_n - 
\underbrace{\displaystyle\frac{\Big|\bx_n\Th\hat\matA\matF^{-1}\hat\matA\Th\bd\Big|^2}{\bd\Th\hat\matA\matF^{-1}\hat\matA\Th\bd}}_\tr{constraint correction},
\end{align}
in which the second term in $\bg_n$ and $z_n$ can be interpreted as a ``constraint correction'', that is only due to the constraint in \eqref{unified.problem}, and may be ignored in the unconstrained case.

Applying \eqref{eq:newton.step.v} to \eqref{eq:newton.update}, we obtain
\begin{align}
\label{eq:newton.step.simple}
\tilde\bu_{\backslash n} = \hat\bu - \frac{\hat{e}_n}{1 - z_n} \, \bg_n.
\end{align}

We are interested in computing the ALO sample prediction error,
\begin{align}
\label{eq:alo.en}
\tilde{e}_n = y_n - \bx_n\Th\tr{vec}\left(\tilde\matW_{\backslash n}\right),
\end{align}
but, for our problem, we first need to reconstruct the parameter $\tilde\matW_{\backslash n}$ from the factors contained in $\tilde\bu_{\backslash n}$. Thus, we have
\begin{aligneq}
\tr{vec}\left(\tilde\matW_{\backslash n}\right) &= \tr{vec}\left[\left(\hat\matU^{(1)} - \matG_n^{(1)} \frac{\hat{e}_n}{1 - z_n}\right)\left(\hat\matU^{(2)} - \matG_n^{(2)} \frac{\hat{e}_n}{1 - z_n}\right)\T\right]\\
&= \hat\bw - \hat\matA\bg_n \frac{\hat{e}_n}{1 - z_n} + \tr{vec}\left(\matG_n^{(1)}\matG_n^{(2)\top}\right) \left(\frac{\hat{e}_n}{1 - z_n}\right)^2,
\end{aligneq}
in which $\matG_n^{(1)} \in \Complex^{M_1 \times R}$ and $\matG_n^{(2)} \in \Complex^{M_2 \times R}$ are matricized slices of $\bg_n$, following the same convention of obtaining $\hat\matU^{(1)}$ and $\hat\matU^{(2)}$ from $\hat\bu$, and we used the fact that [see \eqref{eq:w.A.u}]
\begin{aligneq}
\hat\matA\bg_n &=
\begin{bmatrix}
\hat\matA^{(1)} & \hat\matA^{(2)}
\end{bmatrix}
\begin{bmatrix}
\tr{vec}\left(\matG_n^{(1)}\right)\\
\tr{vec}\left(\matG_n^{(2)}\right)
\end{bmatrix}
\\
&= \hat\matA^{(1)} \tr{vec}\left(\matG_n^{(1)}\right) + \hat\matA^{(2)} \tr{vec}\left(\matG_n^{(2)}\right)\\
&= \tr{vec}\left( \matG_n^{(1)}\hat\matU^{(2)\top} \right) + \tr{vec}\left( \hat\matU^{(1)} \matG_n^{(2)\top} \right).
\end{aligneq}
Then, using \eqref{e.hat.n} and \eqref{zn},
\begin{aligneq}\label{e.tilde.n.final}
\tilde{e}_n &= y_n - \bx_n\Th \hat\bw + \bx_n\Th \hat\matA\bg_n \frac{\hat{e}_n}{1 - z_n} - \bx_n\Th \tr{vec}\left(\matG_n^{(1)}\matG_n^{(2)\top}\right) \left(\frac{\hat{e}_n}{1 - z_n}\right)^2\\
&= \frac{\hat{e}_n}{1 - z_n} - \bx_n\Th \tr{vec}\left(\matG_n^{(1)}\matG_n^{(2)\top}\right) \left(\frac{\hat{e}_n}{1 - z_n}\right)^2\\
&\approx \frac{\hat{e}_n}{1 - z_n},
\end{aligneq}
in which we note that the quadratic term is negligible for sufficiently large $N$. Since $\matF = O(N)$, then $\bg_n = O(1/N)$ and hence $z_n = O(1/N)$ [see \eqref{mat.F}, \eqref{gn}, and \eqref{zn}], while the quadratic term is $O(1/N^2)$, thus negligible for $N \gg (M_1+M_2)R$, as also verified experimentally. This leads to the following ALO metric:
\begin{align}
\label{eq:alo.alpha}
J_{\tr{ALO}}(\alpha) = \frac{1}{N} \sum_{n=1}^N \left|\frac{\hat{e}_n}{1 - z_n}\right|^2.
\end{align}

Note, from \eqref{zn}, that
\begin{align}
z_n = \bx_n\Th\hat\matA\matF^{-1}\hat\matA\Th\bx_n - \underbrace{\displaystyle\frac{\Big|\bx_n\Th\hat\matA\matF^{-1}\hat\matA\Th\bd\Big|^2}{\bd\Th\hat\matA\matF^{-1}\hat\matA\Th\bd}}_\tr{constraint correction},
\end{align}
in which the second term can be interpreted as a ``constraint correction'', that is only due to the constraint in \eqref{unified.problem}, and may be ignored in the unconstrained case.


Note that $\gamma = \sum_{n=1}^N z_n$ may be interpreted as the ``effective number of parameters'' of the solution \cite{Moody1991}, and is the basis of the formulation of the well-known Generalized Cross-Validation (GCV) \cite{Golub1979}. This observation leads to further approximation,
\begin{aligneq}
\label{eq:alo.alpha.trace}
J_{\tr{ALO}}(\alpha)
&\approx \frac{1}{N}\frac{\|\by - \matX\Th \hat\bw\|^2_2}{(1 - \gamma/N)^2},\\
&= \frac{v_y + \hat\bw\Th\matR_{\bx}\hat\bw - 2 \br_{\bx y}\Th\hat\bw}{(1 - \gamma/N)^2},
\end{aligneq}
where $v_y = \|\by\|^2_2 / N$, and a numerically stable and efficient expression of $\gamma$ is shown in \ref{app:zn}. The approximation \eqref{eq:alo.alpha.trace} is preferable to \eqref{eq:alo.alpha} when $N > M$, because we can use $\matR_{\bx}$,  $\br_{\bx y}$ and $v_y$ instead of $\matX$ and $\by$, which is more computationally efficient.
The optimal regularization parameter is then found as
\begin{align}\label{eq:alpha.hat}
\hat\alpha = \argmin_{\alpha} J_{\tr{ALO}}(\alpha),
\end{align}
which can be done using any efficient search.

\section{Experimental results}\label{sec:results}

\subsection{Error-minimization problem: system identification}

In the error-minimization problem, both $\bx(n)$ and $y(n)$ are assumed known, and the filtering error is given by
\begin{align}
\label{linear.model}
e(n) &= y(n) - \bw\Th\bx(n), \quad  n=0,1,\ldots,N-1,
\end{align}
thus the MMSE problem solves 
\begin{align}
\label{hat.h}
\hat\bw &= \argmin_{\bw} \Ex\left[|y(n) - \bw\Th\bx(n)|^2\right],
\end{align}
which corresponds to the unconstrained version of the problem [see \eqref{unified.problem}]. Note that, by setting $\lambda = 0$ in \eqref{eq:ridge}, we obtain the $\ell_2$-regularized solution to the error-minimization problem
\begin{align}\label{err.min.solution}
\hat\bw_{\ell_2} &= (\matR_{\bx} + \alpha \matI)^{-1}{\br}_{\bx y}.
\end{align}

We evaluate the proposed model for the application of system identification, in which we consider
\begin{align}
\label{linear.model.exp}
y(n) &= \bh\T\bx(n) + e(n), \quad  n=0,1,\ldots,N-1,
\end{align}
where the output signal $y(n)$ is obtained by applying the linear filter/weights $\bh \in \Real^M$ to the input signal:
\begin{align}
\bx(n) = [x(n), x(n-1),\dots, x(n-M+1)]\T.
\end{align}
Here, we use $N=1000$, $x(n)$ is simulated from an AR(1) process, \ie $x(n)=ax(n-1)+u(n)$, where $u(n)$ is generated from a zero-mean unit-variance white Gaussian noise and $a=0.9$, and $10$ realizations of the signal and noise are considered.

\begin{figure}[ht!]
\centering
\includegraphics[width=.5\textwidth]{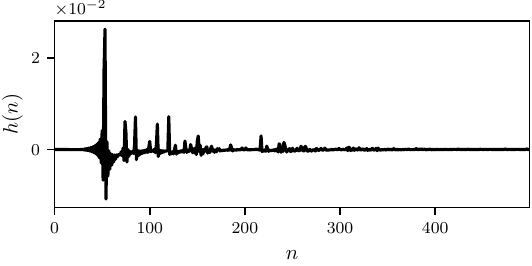}
\caption{Simulated room impulse response considered in the experiment.}
\label{fig:impulses}
\end{figure}

We evaluate an impulse response $\bh$ with length $M=500$ and matrix representation $\matH=\tnr{mat}_{M_1,M2}(\bh) \in \Real^{M_1 \times M_2}$ with dimensions $(M_1, M_2) = (20, 25)$, such that $\tr{rank}(\matH) \leq 20$. The impulse response shown in Fig.~\ref{fig:impulses} is calculated using the room impulse response generator software \cite{Habets2025} for a room of dimensions $(5, 4, 6)$~m, the source in position $(2, 3.5, 2)$~m, the receiver in position $(2, 1.5, 1)$~m, a sampling rate of $8$~kHz, and a reverberation time of $150$~ms.



The quality of the estimation will be assessed using the relative error (or a \emph{misalignment}) of the estimated parameters, defined as
\begin{align}
\label{eq:misalignment}
\mfm(\alpha) &= 10 \log_{10} \left( \frac{\left\|\hat\matW(\alpha) - \matH \right\|^2_{\tr{F}}}{\|\matH\|^2_{\tr{F}}} \right) \quad [\tr{dB}],
\end{align}
where $\hat\matW(\alpha) = \tnr{mat}_{M_1,M2}(\hat\bw)$ is the solution to the unconstrained version of problem \eqref{unified.problem} (\ie with $\bd=\bzero$, $c=0$); when $\hat\matW(\alpha)=\bzero$, a $100\%$ error occurs, corresponding to $\mfm(\alpha) = 0~\tnr{dB}$.


We also denote the rank of the estimated solution as
\begin{align}\label{eq:rank.hat}
\hat{R}(\alpha) = \tr{rank}[\hat\matW(\alpha)].
\end{align}

\begin{figure}[ht]
\centering
\includegraphics[width=\linewidth]{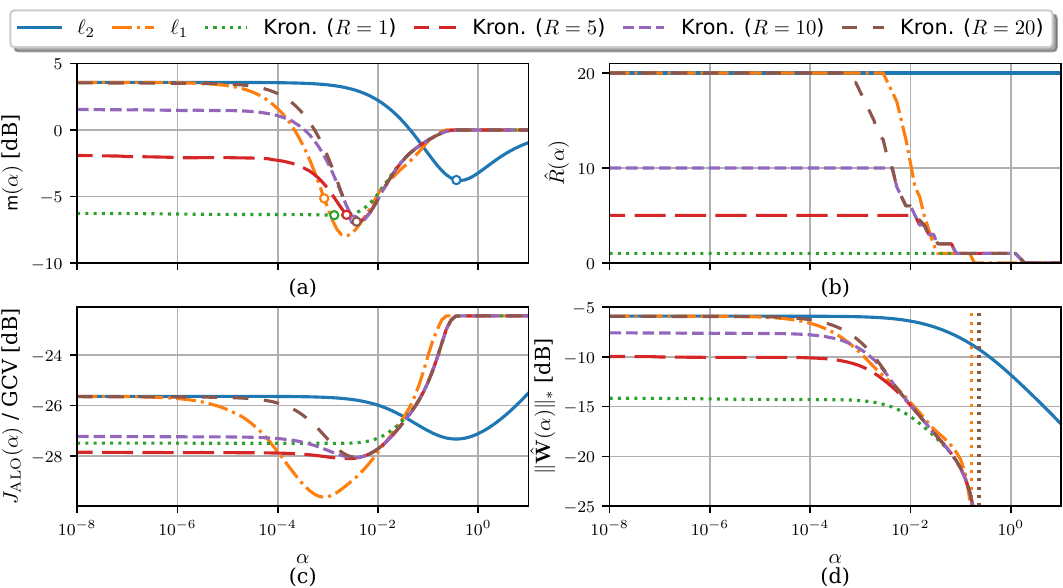}
\caption{(a) Misalignment, (b) rank, (c) ALO/GCV metric, and (d) nuclear norm of solutions obtained by varying $\alpha$ for different regularizers. Solutions are obtained by switching the regularization function in \eqref{unified.problem}, for the unconstrained case ($\bd=\bzero$, $c=0$). To compute the rank and nuclear norm, $\ell_2$ and $\ell_1$ solutions are reshaped to $(20, 25)$. The small circle markers in (a) indicate the misalignment obtained by choosing $\alpha$ from (c). The vertical dotted lines in (d) indicate $\alpha_{\max}$ for $\ell_1$ and nuclear norms, from left to right, respectively [see \eqref{alpha.max}]. Here $N=1000$ and $\tr{SNR}=5$ [dB].}
\label{fig:sensitivity}
\end{figure}

Our goal is to evaluate how the regularization parameter $\alpha$ affects the key properties of the estimate. In Fig.~\ref{fig:sensitivity}(a) and Fig.~\ref{fig:sensitivity}(c), we show, respectively, the misalignment \eqref{eq:misalignment} and, the ALO metric \eqref{eq:alo.alpha} (or GCV for $\ell_1$ and $\ell_2$) as a function of $\alpha$ for different regularizers: $\ell_2$ [corresponding to \eqref{err.min.solution}], $\ell_1$ [corresponding to \eqref{eq:lasso}], and Kronecker product [obtained by solving \eqref{cp.reg.regression} using ALS with different construction ranks $R$].  In Fig.~\ref{fig:sensitivity}(b) and Fig.~\ref{fig:sensitivity}(d), we have, respectively, the corresponding rank and nuclear norm of the solutions. For the estimation via ALS, the number of iterations is set to $I_\tr{ALS} = 100$, which is sufficient for convergence.
We observe the following from Fig.~\ref{fig:sensitivity}. 
\begin{itemize}

\item The Kronecker solutions can have a very similar misalignment to the $\ell_1$ one, both with smaller misalignment than the $\ell_2$ regularized solution. Note that, in Fig.~\ref{fig:sensitivity}~(a), $\ell_1$ outperforms Kronecker for properly chosen $\alpha$. This difference in performance is related to the structure of the underlying impulse response, which is sparse, favoring the $\ell_1$ regularizer (see Fig.~\ref{fig:impulses}), but also composed of a low-rank structure (well-explained by Kronecker).




\item Using $\alpha\approx 0$ (\ie no regularization), we obtain the rank of the solution equal to the construction rank, \ie $\hat{R}(0) = R$. In such a case, a clear improvement (of the misalignment) with respect to the full-rank solution ($R=20$) and the other approaches ($\ell_2$ and $\ell_1$) is obtained by searching for a sufficiently small $R$. In our case, we see in Fig.~\ref{fig:sensitivity}(a) that, for small $\alpha$, $R=10$ produces the best misalignment. This is a common interpretation in the literature: it is necessary to select $R$ to improve performance.

\item As $\alpha$ increases, the nuclear norm of the Kronecker solutions decreases, which also corresponds to low-rank solutions, as seen in Fig.~\ref{fig:sensitivity}(c) and Fig.~\ref{fig:sensitivity}(b), respectively. This illustrates well the equivalency between \eqref{cp.nuclear.reg} and \eqref{cp.reg.regression}, as well as the expected rank reduction [seen in Fig.~\ref{fig:sensitivity}(b)] and the existence of $\alpha_\tr{max}$ [the second vertical line in Fig.~\ref{fig:sensitivity}(c)]. Note that $\ell_1$ estimates follows a similar reduction in nuclear norm, since as the number of zero elements increases (induced by the $\ell_1$ norm), the higher the chance of zeroing a full column/line of the matricized solution, which leads to a low-rank structure. On the other hand, in the $\ell_2$ solution some reduction of the nuclear norm occurs for large $\alpha$, but it does not translate into a reduction of the rank because there is no constraint on the structure of the solution.


\item The optimal regularization parameters $\hat\alpha$, shown as circle markers in Fig.~\ref{fig:sensitivity}~(a), are obtained from ALO \eqref{eq:alo.alpha.trace} for the Kronecker solutions and lead to close-to-optimal misalignment performance. This confirms the efficacy of the ALO metric. In fact, it may even outperform $\ell_1$ (and clearly $\ell_2$) solutions obtained from GCV, even though $\ell_1$ theoretically can have a lower misalignment. This theoretical advantage is due to the form of the room impulse response, which is sparse (see Fig.~\ref{fig:impulses}); the Kronecker structure is related to the low-rank property, which is also present, but is less prominent.

\item Note that Kronecker solutions $\hat\matW(\alpha)$ share a very similar $\hat\alpha$. In fact, for sufficiently high construction rank $R$, the Kronecker curves in Fig.~\ref{fig:sensitivity} converge as $\alpha$ increases from $\hat\alpha$, essentially leading to the same solution, since their rank $\hat{R}(\alpha)$ becomes similar. This fact confirms our observation in the end of Sec.~\ref{sec:lowrank} that we can use a large $R$ and yet, with the suitable chosen regularization parameter $\alpha$, we can obtain the optimal nuclear norm and rank.




\end{itemize}

\subsection{Interference suppression problem: beamforming}\label{Sec:Numerical.MVDR}

In the interference suppression problem, only $\bx(n)$ is known, and the goal is to minimize the (energy of) interference in the filtered output $\bw\Th\bx(n)$, while maintaining the energy of the desired signal, as enforced by the constraint $\bw\Th\bd = 1$,
\begin{align}
\label{interference.problem}
\hat\bw & = \argmin_{\bw}  \Ex\left[|\bw\Th\bx(n)|^2\right] \quad \tnr{s. t. }\quad \bd\Th\bw=1,
\end{align}
known, in the context of array processing as a minimum variance distortionless response (MVDR) beamformer. The solution for the $\ell_2$-regularized problem is obtained by
\begin{align}\label{int.sup.solution}
\hat\bw_{\ell_2} &= \frac{(\matR_{\bx} + \alpha \matI)^{-1}\bd}{\bd\Th(\matR_{\bx} + \alpha \matI)^{-1}\bd},
\end{align}
where we set $c=1$ and $\by = \bzero$ in \eqref{eq:ridge} (and, consequently, $\br_{\bx y} = \bzero$).

In this context, we evaluate the proposed model for the application of antenna beamforming, in which we consider
\begin{align}
\bx(n) = \sum_{k=0}^K \bs_k(n) + \be(n),
\end{align}
where $\be(n)$ is a zero-mean, circular complex Gaussian noise with covariance matrix $\Ex[\be(n)\be\Th(n)]=\matI$, and $\bs_k(n)$ is the baseband signal vector corresponding to the $k$-th user. Assuming $L$ multipath components,
\begin{align}
\bs_k(n) = \sigma_k \sum_{l=1}^L \beta_{k,l}(n)\bd(\phi_{k,l}),
\end{align}
where $\sigma_k^2$ is the power of user $k$, $\beta_{k,l}(n)$ models the $l$-th path of the $k$-th user, and $\bd(\phi_{k,l})$ is the steering vector of the $l$-th path from the $k$-th user with angle of arrival $\phi_{k,l}$. The steering vector is defined as
\begin{align}
\bd(\phi) = \bd_2(\phi) \otimes \bd_1(\phi)
\end{align}
where
\begin{align}
\bd_1(\phi) &= [1, \e^{-j \pi \cos(\phi)}, \e^{-j 2\pi \cos(\phi)},\dots, \e^{-j(M_1-1) \pi \cos(\phi)}]\T,\\
\bd_2(\phi) &= [1, \e^{-j \pi \sin(\phi)}, \e^{-j 2\pi \sin(\phi)},\dots, \e^{-j(M_2-1) \pi \sin(\phi)}]\T,
\end{align}
that is, we assume that $\bx(n)$ is acquired at a uniform rectangular antenna array with $M = M_1 \times M_2$ elements spaced at half-wavelength, following the convention of \cite[Ch.~6.1]{Benesty2019}.

The true covariance matrix is thus calculated as
\begin{align}
\ov\matR_{\bx} &= \sum_{k=0}^K \sigma_k^2 \matD_k + \matI,
\end{align}
where
\begin{align}
\matD_k &= \frac{1}{L}\sum_{l=1}^L  \bd(\phi_{k,l})\bd\Th(\phi_{k,l}).
\end{align}

The quality of interference suppression is measured by the \gls{sinr} at the output of the filter, calculated as
\begin{align}
\label{SINR.opt}
\mf{SINR} &=  \frac{\hat\bw\Th\matR_\tr{soi}\hat\bw}
{\hat\bw\Th\matR_\tr{int}\hat\bw},
\end{align}
where the covariance matrices $\matR_\tr{soi} = \sigma_0 \matD_0$ and $\matR_\tr{int} = \ov\matR_{\bx} - \matR_\tr{soi}$, in which we consider the signal-of-interest (SOI) at index $0$.

In this example, we have $M_1=M_2=4$, and $L=12$ fading paths with spread of $\Delta\phi_k = 5^{\circ}$ around the mean angle-of-arrival, $K=6$ interference signals located at $\bphi = [-1/3, -5/18, -1/6, 0, 2/9, 7/18]\, \pi$, with the SOI at $\phi_0 = \pi/6$.


We want to evaluate the $\mf{SINR}$ as a function of number of snapshots $N$ for different rank: $R=1$, the minimum rank, and $R=4$, which is the maximum possible rank in this example. We average the $\mf{SINR}$ over $100$ realizations of the signal and noise. The $\ell_2$ solution is compared to the Kronecker-product solutions, as well as the optimal and matched filters, where we use different approaches to select the regularization parameter $\alpha$: a) ``oracle'': $\alpha$ is set to produce the highest $\mf{SINR}$, b) ALO: $\alpha$ obtained by minimizing \eqref{eq:alo.alpha.trace}, c) AIC: $\alpha$ estimated using the approach of \cite{Zhou2013a}\footnote{We omitted estimations made with the BIC, as they were too similar to AIC.} with the degrees of freedom $\gamma$ computed from \ref{app:zn}. Full-rank solution of \eqref{int.sup.solution} is treated as the comparison baseline, and we use GCV for the $\alpha$ search.


\begin{figure}[ht!]
\centering
\includegraphics[width=0.5\textwidth]{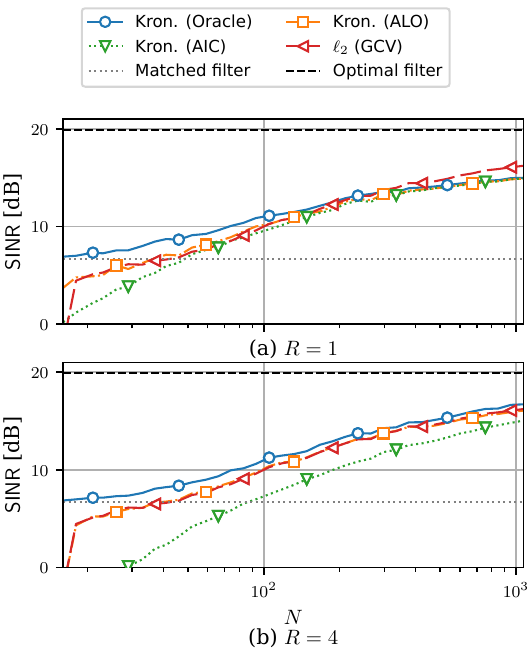}
\caption{SINR as a function of number of snapshots $N$ for different methods, with construction rank (a) $R=1$ and (b) $R=4$.}
\label{fig:mis.rank}
\end{figure}

\begin{figure}[ht!]
\centering
\includegraphics[width=0.5\linewidth]{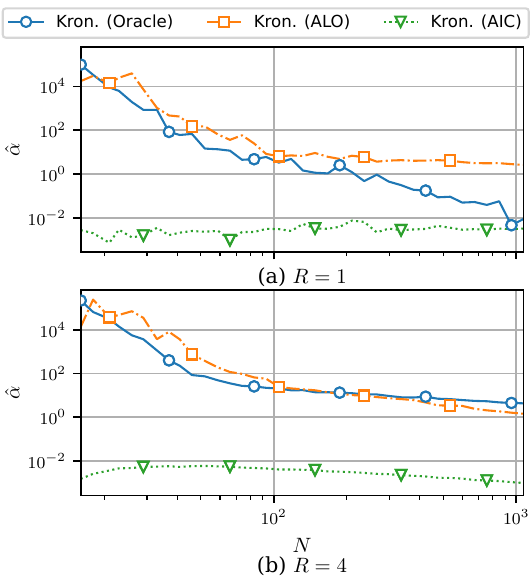}
\caption{Estimated $\alpha$ as a function of number of snapshots $N$ for different methods, with construction rank (a) $R=1$ and (b) $R=4$.}
\label{fig:alpha.rank}
\end{figure}

In Fig.~\ref{fig:mis.rank}, we show the average SINR of the considered scenarios and, in Fig.~\ref{fig:alpha.rank}, the corresponding estimated $\alpha$. 
We make the following observations.
\begin{itemize}


\item Inadequate regularization can potentially lead to very poor performance that can be worse than the input signal (which has $0$ dB SINR), especially when only a small number of snapshots $N$ is available. In such a case, mitigating this problem usually involves selecting $R$, but, even if $R=1$ is chosen, there is still an improvement in SINR by estimating $\alpha$ properly.



\item Regularization based on the ALO approach closely matches the $\ell_2$ solution in the full-rank case of $R=4$ [see Fig.~\ref{fig:mis.rank}~b)], and for $R=1$ when $N$ is small [see Fig.~\ref{fig:mis.rank}~a)]. The advantage here is obtaining full-rank performance with a low-rank ($R=1$) model, which has less parameters and is more computationally efficient.


\item The ALO method outperforms the AIC in general. In fact, selecting $\alpha$ from the AIC attains the worst SINR, which can be explained by its poor $\alpha$ estimation, in Figure~\ref{fig:alpha.rank}. The performance of ALO is close to the oracle, both for $R=1$ and $R=4$, which is due to the close $\alpha$ estimates produced for small to moderate $N$.




\end{itemize}

\section{Conclusion}\label{sec:conclusion}

In this work, we presented a method to estimate the regularization parameter in the problem of low-rank MMSE filter estimation with a Kronecker-product representation. 
We explore how the regularization of the factors in the representation is connected to the problem of rank minimization, and how a practical approach to such a problem can be the estimation of the regularization parameter. 
Numerical simulations indicate that the proposed approximate leave-one-out (ALO) approach performs very close to the oracle solution, providing significant improvements in performance compared to other methods commonly used in the literature.
We also note that the low-rank model tends to have greater improvements in scenarios in which the data is noisy and/or scarce, guiding future work towards problems in which the available data has such characteristics.

\section*{Acknowledgments}

This work was supported in part by the \textit{Fonds de recherche du Québec} (FRQ) - \textit{Nature et technologies} under the \textit{Doctoral research scholarships B2X 2024-2025} program, file number 342496, recipient Daniel~Gomes~de~Pinho~Zanco.

During the preparation of this work, the author(s) used Claude for aiding in research and coding. The author(s) reviewed and edited the output as needed and take full responsibility for the content of the published article. 

\appendix
\section{Definition of $\matA^{(1)}$ and $\matA^{(2)}$}
\label{app:factors}

We have, from \eqref{cp.w}, that
\begin{align}
\bw &= \sum_{r=1}^R \bw_r,
\end{align}
where $\bw_r$'s are rank-$1$ terms of the representation, and may be expressed as
\begin{equation}
\begin{aligned}
\bw_r &= \bu_{r}^{(2)} \otimes \bu_{r}^{(1)}\\
&= \left(\bu_{r}^{(2)} \otimes \matI_{M_1}\right) \bu_{r}^{(1)}\\
&= \left(\matI_{M_2} \otimes \bu_{r}^{(1)}\right) \bu_{r}^{(2)}.
\end{aligned}
\end{equation}
Thus, we may express $\bw$ as
\begin{align}
\label{eq:w.Ar.ur}
\bw 
&= \sum_{r=1}^R \matA_{r}^{(k)} \bu_{r}^{(k)}, \quad k=1,2,
\end{align}
where
\begin{align}
\matA_{r}^{(1)} &= \bu_{r}^{(2)} \otimes \matI_{M_1},\\
\matA_{r}^{(2)} &= \matI_{M_2} \otimes \bu_{r}^{(1)}.
\end{align}
Furthermore, we may also express \eqref{eq:w.Ar.ur} as
\begin{align}\label{eq:w.A.u}
\bw &= \matA^{(k)} \bu^{(k)}, \quad k=1,2,
\end{align}
where
\begin{align}
\matA^{(k)} &=
\begin{bmatrix}
\matA^{(k)}_1 & \matA^{(k)}_2 & \cdots & \matA^{(k)}_R
\end{bmatrix} \in \Complex^{M \times M_k R} ,\quad k=1,2,
\end{align}
and
\begin{equation}
\begin{aligned}
\bu^{(k)} &= \tr{vec}\left(\matU^{(k)}\right)\\
&=
\begin{bmatrix}
\bu^{(k) \top}_1 & \bu^{(k) \top}_2 & \cdots & \bu^{(k) \top}_R
\end{bmatrix}\T  \in \Complex^{M_k R} , \quad k=1,2.
\end{aligned}
\end{equation}

We may view the matrices $\matA^{(1)}$ and $\matA^{(2)}$ as functions of $\matU^{(2)}$ and $\matU^{(1)}$, respectively, thus we have
\begin{align}
\label{eq:A.1}
\matA^{(1)} \equiv \matA^{(1)}\left(\matU^{(2)}\right) &= \matU^{(2)} \otimes \matI_{M_1},\\
\label{eq:A.2}
\matA^{(2)} \equiv \matA^{(2)}\left(\matU^{(1)}\right) &=  \left(\matI_{M_2} \otimes \matU^{(1)}\right) \matK_{M_2,R},
\end{align}
where $\matK_{M_2,R}$ is a commutation matrix, which reorders $\bu^{(2)} = \tr{vec}(\matU^{(2)})$ into $\tr{vec}(\matU^{(2) \top})$.

\section[Linearization of $C(\cdot)$ and derivation of the derivatives of $L(\cdot)$]{Linearization of $C(\bu)$ and derivation of the derivatives of $L_{\backslash_n}(\bu)$}
\label{app:alo}

We note that
\begin{aligneq}
\label{eq:border.pre}
C(\bu + \hat\bu) 
&= \frac{1}{2} \bd\Th\matA(\bu + \hat\bu)(\bu + \hat\bu) - c\\
&= \underbrace{\frac{1}{2} \bd\Th\matA(\hat\bu) - c}_{0} + \frac{1}{2}\bd\Th\matA(\hat\bu)\bu + \frac{1}{2}\bd\Th\matA(\bu)\hat\bu + \frac{1}{2} \bd\Th\matA(\bu)\bu\\
&= \bd\Th\matA(\hat\bu)\bu + \frac{1}{2} \bd\Th\matA(\bu)\bu,
\end{aligneq}
where we use the facts that $\hat\bu$ respects the constraint, \ie $C(\hat\bu) = 0$, $\matA(\bu + \hat\bu) = \matA(\bu) + \matA(\hat\bu)$, and $\matA(\hat\bu)\bu = \matA(\bu)\hat\bu$. Thus the constraint in \eqref{eq:newton.step} may be linearized as
\begin{align}
C(\bu + \hat\bu) \approx \bd\Th\matA(\hat\bu)\bu.
\end{align}





The gradient is given by
\begin{align}\label{eq:grad.full}
\nabla_{\bu^*} L_{\backslash n}(\bu) &= \nabla_{\bu^*} L(\bu) - \nabla_{\bu^*} \ell_n(\bu),
\end{align}
where we perform complex differentiation using the convention from \cite[Appendix~B.2]{Haykin2014}
\begin{aligneq}\label{eq:grad.n.fun}
\nabla_{\bu^*} \ell_n(\bu) 
&= \frac{\dd}{\dd \bu\Th} \ell_n(\bu)\\
&= \frac{\dd}{\dd \bu\Th} \left(\left| y_n - \frac{1}{2}\bx_n\Th\matA\bu \right|^2\right)\\
&= -\frac{1}{2}\frac{\dd}{\dd \bu\Th}\left(\bu\Th\matA\Th\right)\bx_n \left( y_n - \frac{1}{2}\bx_n\Th\matA\bu \right)\\
&= -\matA\Th\bx_n\left( y_n - \frac{1}{2}\bx_n\Th\matA\bu \right) ,
\end{aligneq}
in which, using the fact that $\matA\bu = \matA^{(1)}\bu^{(1)} + \matA^{(2)}\bu^{(2)} = 2\matA^{(1)}\bu^{(1)} = 2\matA^{(2)}\bu^{(2)}$,
\begin{align}
\label{eq:nabla.Au}
\frac{\dd}{\dd \bu\Th}\left(\bu\Th\matA\Th\right) 
= 2
\begin{bmatrix}
\frac{\dd}{\dd \bu^{(1)\mr{H}}} \matA^{(1)\mr{H}}\bu^{(1)\mr{H}}\\
\frac{\dd}{\dd \bu^{(2)\mr{H}}} \matA^{(2)\mr{H}}\bu^{(2)\mr{H}}
\end{bmatrix} 
= 2 \matA\Th.
\end{align}

Then, from \eqref{eq:obj.fun} and \eqref{eq:grad.n.fun}, we have
\begin{aligneq}
\nabla_{\bu^*} L(\bu) &= N\alpha \bu -\sum_{n=1}^N \matA\Th\bx_n\left( y_n - \frac{1}{2}\bx_n\Th\matA\bu \right)\\
&= N\alpha \bu + \frac{N}{2}\matA\Th\matR_{\bx}\matA\bu - N\matA\Th\br_{\bx y}.
\end{aligneq}


To calculate the second order terms, we start by differentiating \eqref{eq:grad.n.fun} again, yielding
\begin{aligneq}
\label{eq:hess.alo.func}
\nabla^2_{\bu^*} \ell_n(\bu) 
&= \frac{\dd}{\dd\bu}\left[-\matA\Th\bx_n\left( y_n - \frac{1}{2}\bx_n\Th\matA\bu \right)\right]\\
&= -\matA\Th\bx_n\frac{\dd}{\dd\bu}\left( y_n - \frac{1}{2}\bx_n\Th\matA\bu \right)\\
&= \matA\Th\bx_n\bx_n\Th\matA,
\end{aligneq}

and
\begin{aligneq}
\label{eq:hess.real.alo.func}
\nabla^2_{\bu} \ell_n(\bu) 
&= \frac{\dd}{\dd\bu}\left[-\matA\T\bx_n^*\left( y_n - \frac{1}{2}\bx_n\Th\matA\bu \right)^*\right]\\
&= -\matB_n,
\end{aligneq}
in which 
\begin{align}
\matB_n = \left( y_n^* - \frac{1}{2}\matA\Th\bu\Th\bx_n^* \right) \nabla_{\bu}\left(\matA\T\bx_n^*\right) 
\end{align} 
and
\begin{align}
\label{eq:nabla.A.xn}
\nabla_{\bu}\left(\matA\T\bx_n\right) = 
\begin{bmatrix}
\nabla_{\bu^{(1)}}\matA^{(1) \top}\bx_n & \nabla_{\bu^{(1)}}\matA^{(2) \top}\bx_n\\
\nabla_{\bu^{(2)}}\matA^{(1) \top}\bx_n & \nabla_{\bu^{(2)}}\matA^{(2) \top}\bx_n\\
\end{bmatrix},
\end{align}
where
\begin{align}
\nabla_{\bu^{(k)}}\matA^{(k) \top}\bx_n^* &= \bzero \in \Complex^{M_kR \times M_kR}, \quad k=1,2,\\
\label{eq:nabla.u1.A2}
\nabla_{\bu^{(1)}}\matA^{(2) \top}\bx_n^* &= \matI_R \otimes \tr{mat}_{M_1, M_2}(\bx_n^*)  \in \Complex^{M_1R \times M_2R},\\
\label{eq:nabla.u2.A1}
\nabla_{\bu^{(2)}}\matA^{(1) \top}\bx_n^* &= \matI_R \otimes \tr{mat}_{M_1, M_2}(\bx_n^*)\T \in \Complex^{M_2R \times M_1R}.
\end{align}

Finally, we have
\begin{align}\label{eq:hess.full}
\nabla^2_{\bu^*} L_{\backslash n}(\bu) &= \sum_{i \neq n}^N \matA\bx_i\bx_i\Th\matA + N\alpha\matI,\\
\nabla^2_{\bu} L_{\backslash n}(\bu) &= -\sum_{i \neq n}^N \matB_i.
\end{align}
\section{Derivation of $\gamma$}
\label{app:zn}

Note that $z_n$ can be obtained from the diagonal elements of the matrix
\begin{align}
\matZ = \matX\Th\left( \hat\matA \matF^{-1} \hat\matA\Th - \displaystyle\frac{\hat\matA\matF^{-1}\hat\matA\Th\bd\bd\Th\hat\matA\matF^{-1}\hat\matA\Th}{\bd\Th\hat\matA\matF^{-1}\hat\matA\Th\bd} \right)\matX,
\end{align}
and taking the trace, yields the effective number of parameters
\begin{aligneq}\label{gamma.appendix}
\gamma = \tr{tr}(\matZ) 
&= \tr{tr}\left( N\hat\matA\Th\matR_{\bx}\hat\matA \matF^{-1} \right) - N\displaystyle\frac{\bd\Th\hat\matA\matF^{-1}\hat\matA\Th\matR_{\bx}\hat\matA\matF^{-1}\hat\matA\Th\bd}{\bd\Th\hat\matA\matF^{-1}\hat\matA\Th\bd}\\
&= \tr{tr}\left( \matI - N\alpha \matF^{-1} \right) - \displaystyle\frac{\bd\Th\hat\matA\matF^{-1}(\matI - N\alpha \matF^{-1})\hat\matA\Th\bd}{\bd\Th\hat\matA\matF^{-1}\hat\matA\Th\bd}\\
&= (M_1+M_2)R - 1 - N\alpha\left(\tr{tr}\left(\matF^{-1} \right) - \displaystyle\frac{\bd\Th\hat\matA\matF^{-2}\hat\matA\Th\bd}{\bd\Th\hat\matA\matF^{-1}\hat\matA\Th\bd}\right),
\end{aligneq}
where we used the fact that $N\hat\matA\Th\matR_{\bx}\hat\matA\matF^{-1} = \matI - N\alpha \matF^{-1}$.

Note that we can view $\gamma$ as an unconstrained term minus a term due to the constraint:
\begin{align}\label{gamma.terms}
\gamma = \gamma' - \kappa
\end{align}
where
\begin{align}
\gamma' &= (M_1+M_2)R - N\alpha\tr{tr}\left(\matF^{-1} \right),\\
\kappa &= 1 - N\alpha \displaystyle\frac{\bd\Th\hat\matA\matF^{-2}\hat\matA\Th\bd}{\bd\Th\hat\matA\matF^{-1}\hat\matA\Th\bd}.
\end{align}
Note that, when we are dealing with the unconstrained version of the problem \eqref{unified.problem}, then $\gamma = \gamma'$.

Now, taking the eigenvalue decomposition
\begin{align}
\hat\matA\Th\matR_{\bx}\hat\matA = \matQ \diag\blambda \matQ\Th,
\end{align}
where $\matQ$ is the matrix of eigenvectors and $\blambda = [\lambda_1, \ld, \lambda_P]\T$ are the corresponding eigenvalues (ordered non-increasingly), with $P = (M_1+M_2)R$, we may express 
\begin{align}
\label{gamma.eig}
\gamma' &= P - \sum_{p=1}^P \frac{\alpha}{\lambda_p + \alpha},\\
\label{kappa.eig}
\kappa &= 1 - \frac{\displaystyle\sum_{p=1}^P \frac{\alpha |b_p|^2}{(\lambda_p + \alpha)^2}}{\displaystyle\sum_{p=1}^P \frac{|b_p|^2}{\lambda_p + \alpha}},
\end{align}
where $\bb = \matQ\Th \hat\matA\Th \bd = [b_1, \ld, b_P]\T$.


We know that
\begin{aligneq}
\|\blambda\|_0 &= \tr{rank}\left( \hat\matA\Th\matR_{\bx}\hat\matA \right)\\
&\leq \min\left\{\tr{rank}(\hat\matA), \tr{rank}(\matR_{\bx})\right\}.
\end{aligneq}
Recall that $\matR_{\bx}$ is the $N$-samples estimate of the covariance matrix, so $\tr{rank}(\matR_{\bx}) = \min\{M, N\}$, and $\hat\matA = [\hat\matA^{(1)} \; \hat\matA^{(2)}]$ is a block matrix, so from
\cite[Fact~2.11.8]{Bernstein2009}, we have
\begin{align}
\tr{rank}(\hat\matA) = \tr{rank}\left(\hat\matA^{(1)}\right) + \tr{rank}\left(\hat\matA^{(2)}\right)  - \tr{dim}\left[\tr{range}\left(\hat\matA^{(1)}\right) \cap \tr{range}\left(\hat\matA^{(2)}\right)\right]
\end{align}
where, assuming $R \leq \min\{M_1, M_2\}$,
\begin{align}
\tr{rank}\left(\hat\matA^{(k)}\right) = M_k R, \quad k=1,2,
\end{align}
and the intersection term, using \cite[Fact~7.4.22]{Bernstein2009}, \eqref{eq:A.1}, and \eqref{eq:A.2},
\begin{aligneq}
\tr{dim}\left[\tr{range}\left(\hat\matA^{(1)}\right) \cap \tr{range}\left(\hat\matA^{(2)}\right)\right] &= \tr{dim}\left[\tr{range}\left(\hat\matU^{(2)} \otimes \matI_{M_1}\right) \cap \tr{range}\left(\matI_{M_2} \otimes \hat\matU^{(1)}\right)\right]\\
&= \tr{dim}\left[\tr{range}\left(\hat\matU^{(2)} \otimes \hat\matU^{(1)}\right)\right] = R^2,
\end{aligneq}
thus
\begin{align}
\tr{rank}(\hat\matA) &= (M_1 + M_2)R - R^2 = P - R^2,
\end{align}
and, since $P - R^2 \leq M$ for $R \leq \min \{M_1, M_2\}$,
\begin{align}
\|\blambda\|_0 &\leq \min\left\{P - R^2, N\right\}.
\end{align}

In order to avoid numerical issues when computing $\gamma$ [from \eqref{gamma.eig} and \eqref{kappa.eig}] for small $\alpha > 0$, it is preferable to explicitly zero-out some elements of $\blambda$, \ie $\lambda_p \leftarrow 0$ for $p > \min\left\{P - R^2, N\right\}$, since these eigenvalues are exactly zero in theory.







\bibliography{Common.Files.Bib/IEEEabrv,references.bib}
\bibliographystyle{Common.Files.Bib/IEEEtran}

\end{document}